\documentclass[10pt,twocolumn,letterpaper]{article}

\usepackage{cvpr}              

\usepackage{graphicx}
\usepackage{amsmath}
\usepackage{amssymb}
\usepackage{booktabs}

\usepackage[square,numbers]{natbib}
\usepackage{times}
\usepackage{epsfig}
\usepackage{graphicx}
\usepackage{amsmath}
\usepackage{amssymb}
\usepackage[utf8]{inputenc}
\usepackage{multirow}
\usepackage{booktabs} 
\usepackage{url}
\usepackage{graphicx}
\usepackage{listings} 
\usepackage{xcolor}
\usepackage{floatrow}
\usepackage{siunitx}

\usepackage{paralist}
\usepackage{multirow}
\usepackage{graphicx}
\usepackage[utf8]{inputenc} 
\usepackage[T1]{fontenc}    
\usepackage{url}            
\usepackage{booktabs}       
\usepackage{amsfonts}       
\usepackage{nicefrac}       
\usepackage{microtype}      
\usepackage{amsmath}
\usepackage{color}
\usepackage{amsmath}
\usepackage[ruled,vlined]{algorithm2e}
\usepackage{graphicx}
\usepackage{enumitem}
\usepackage{cancel}
\usepackage[accsupp]{axessibility}
\usepackage[pagebackref,breaklinks,colorlinks]{hyperref}

\usepackage[capitalize]{cleveref}
\crefname{section}{Sec.}{Secs.}
\Crefname{section}{Section}{Sections}
\Crefname{table}{Table}{Tables}
\crefname{table}{Tab.}{Tabs.}

\def\cvprPaperID{7747} 
\def\confName{CVPR}
\def\confYear{2022}

\title{Modular Action Concept Grounding in Semantic Video Prediction}

\author{%
Wei Yu\textsuperscript{\normalfont 1,2},%
 Wenxin Chen\textsuperscript{\normalfont 1,2}, %
Songheng Yin\textsuperscript{\normalfont 1,2}, %
Steve Easterbrook\textsuperscript{\normalfont 1}, %
Animesh Garg\textsuperscript{\normalfont 1,2,3}\\
\textsuperscript{1}University of Toronto, %
\textsuperscript{2}Vector Institute, %
\textsuperscript{3}Nvidia %
}

\begin{document}
\maketitle

\begin{abstract}
Recent works in video prediction have mainly focused on passive forecasting and low-level action-conditional prediction, which sidesteps the learning of interaction between agents and objects. We introduce the task of semantic action-conditional video prediction, which uses semantic action labels to describe those interactions and can be regarded as an inverse problem of action recognition. The challenge of this new task primarily lies in how to effectively inform the model of semantic action information. Inspired by the idea of Mixture of Experts, we embody each abstract label by a structured combination of various visual concept learners and propose a novel video prediction model, \textbf{M}odular \textbf{A}ction \textbf{C}oncept Network (MAC).
Our method is evaluated on two newly designed synthetic datasets, CLEVR-Building-Blocks and Sapien-Kitchen, and 
one real-world dataset called Tower-Creation. Extensive experiments demonstrate that MAC can correctly condition on given instructions and generate corresponding future frames without need of bounding boxes. We further show that the trained model can make out-of-distribution generalization, be quickly adapted to new object categories and exploit its learnt features for object detection, showing the progression 
towards higher-level cognitive abilities. More visualizations can be found at \url{http://www.pair.toronto.edu/mac/}.

\end{abstract}

\section{Introduction}
Recently, video prediction has drawn a lot of attention due to its ability to capture meaningful representations through self-supervision~\cite{wang2018eidetic,yu2019efficient}. Although modern video prediction methods have made significant progress in improving predictive accuracy, most of their applications are limited in the scenarios of passive forecasting~\cite{villegas2017decomposing,wang2018predrnn++,byeon2018contextvp,jin2020exploring}, meaning models can only passively observe a short period of dynamics and accordingly make a short-term extrapolation. Such settings neglect the fact that the observer can also become an active participant in the environment. 

To model the movements of active manipulators, several low-level action-conditional video prediction models have been proposed in the community~\cite{oh2015action,mathieu2015deep,babaeizadeh2017stochastic,ebert2017self,menapace2021playable,huang2021layered}. In this work, we go one step further by introducing the task of \textit{semantic action-conditional video prediction} which emphasizes the modeling of interactions between agents and environment. Instead of using low-level single-entity actions such as action vectors of robot arms as done in prior works~\cite{finn2016unsupervised,kurutach2018learning}, our new task provides semantic descriptions of interactive actions, e.g. \textit{"Open the door"}, and asks the model to imagine \textit{"What if I open the door"} in the form of future frames. This task requires the model to recognize the object identity, assign correct affordances to objects and envision the long-term expectation by planning a reasonable trajectory toward the goal, which resembles how humans might imagine conditional futures. The ability to predict correct and semantically consistent future perceptual information is indicative of conceptual grounding of actions, in a manner similar to object grounding in image-based detection and generation tasks.


\begin{figure}[!t]
	\centering
	\includegraphics[width=1\textwidth]{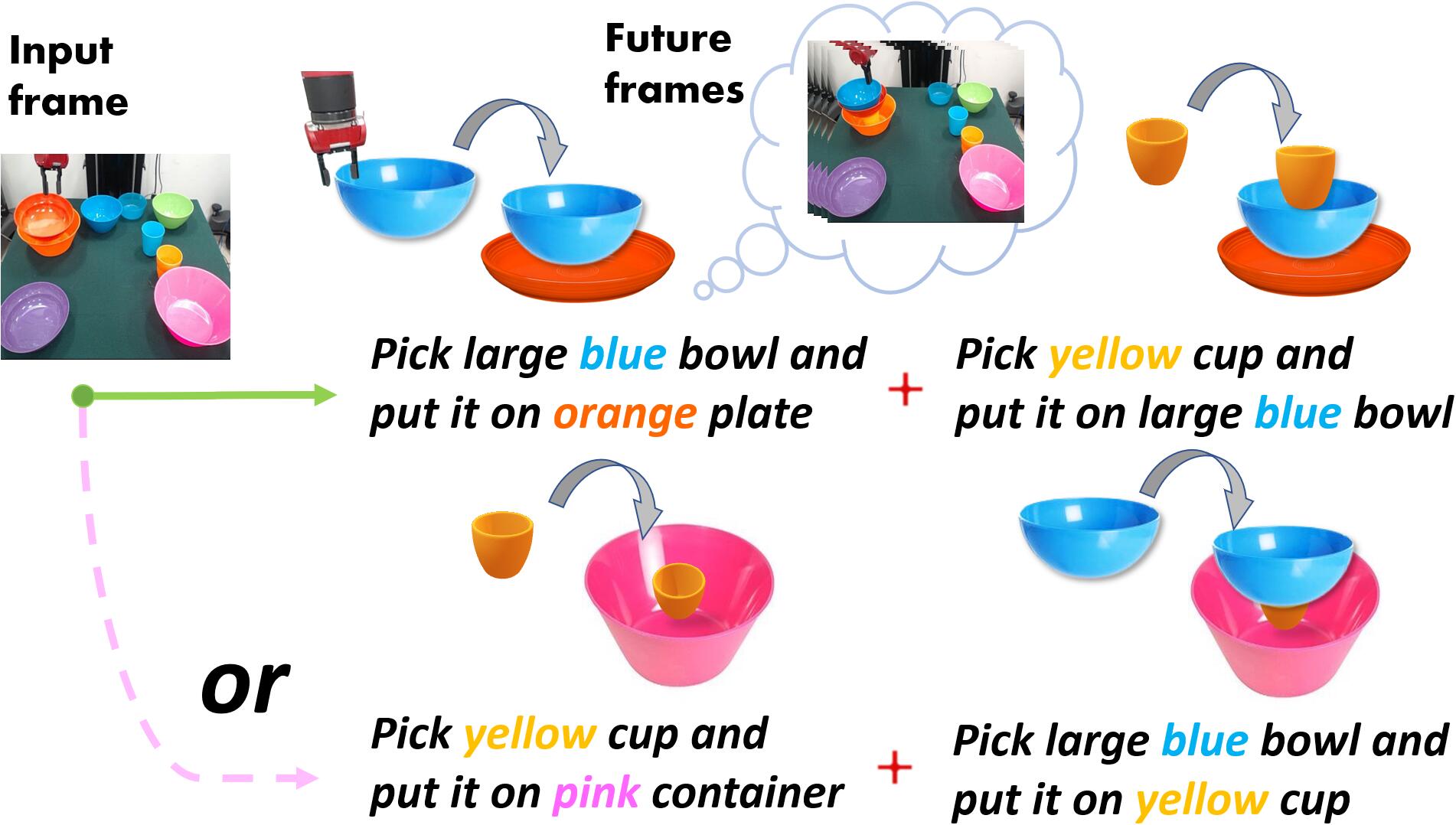}
	\caption{\textbf{Concept Grounding in Semantic Video Prediction}. After observing the scene, an agent predicts future frames conditioned on a series of semantic actions describing agent-object interactions. Neither bounding boxes nor key points are provided. Conditioning on different action labels leads to \textit{Counterfactual} generations.}
		\label{figure:task}
\end{figure}
The challenge of action-conditional video prediction primarily lies in how to correctly inform the model of more abstract semantic action information. Existing low-level counterparts usually achieve this by employing a naive concatenation~\cite{finn2016unsupervised,babaeizadeh2017stochastic} with action vector of each timestep. While this implementation might enable model to move the desired objects, it fails to produce consistent long-term  predictions toward target locations in the multi-entity settings because it was originally designed to only encode the motion information of a single entity. If we take "\textit{put A on B}" as an example, it turns out to be difficult to make the model learn what and where \textit{B} is, because the 
main self-supervisory signals in the 
framework of video prediction are pixel changes and \textit{B} is not moving in this case. In order to distinguish and locate instances in the scene, other related works heavily rely on pre-trained object detectors or ground-truth bounding boxes~\cite{bar2020compositional,ji2020action,huang2018finding,wu2020future}. However, we argue that utilizing a pre-trained detector actually simplifies the task since such a detector already solves the major difficulty by mapping high-dimension inputs to low-dimension groundings. Furthermore, bounding boxes cannot effectively describe complex visual changes including rotations and occlusions. Thus, a more flexible way of representing objects and actions is required.

We present a new video prediction model, MAC, short for Modular Action Concept Network. Inspired by the idea of Mixture of Experts, MAC embodies each semantic label by a structured combination of various concept slots, each of which encodes the spatial representation of a specific concept. This design allows MAC to reuse and integrate the knowledge learnt from different scenarios so that it can perceive the locations of motionless objects and extrapolate to unseen cases, showing the progression towards higher-level cognitive abilities.
The contributions of this work are summarized as follows:
 
    
\begin{enumerate}[
    itemsep=0ex,
    topsep=1pt,
    leftmargin=*
    ]
\item We introduce a new task, semantic action-conditional video prediction as illustrated in Fig \ref{figure:task}, which can be viewed as an inverse problem of action recognition.

\item We create two new synthetic video datasets, CLEVR-Building-blocks and Sapien-Kitchen, and label one real-world dataset called Tower-Creation  for evaluation.

\item We propose a novel video prediction model, Modular Action Concept Network, in which aggregation of visual concept slots is directly controlled by action labels. We show that MAC can successfully depict the long-term counterfactual evolution without need of bounding boxes. 
\item We demonstrate that the trained MAC can make out-of-distribution generalization, be adapted for new object categories with a small number of samples and exploit its learnt features for detection.
\end{enumerate}
\begin{figure*}
	\centering
		\includegraphics[width=1.0\textwidth]{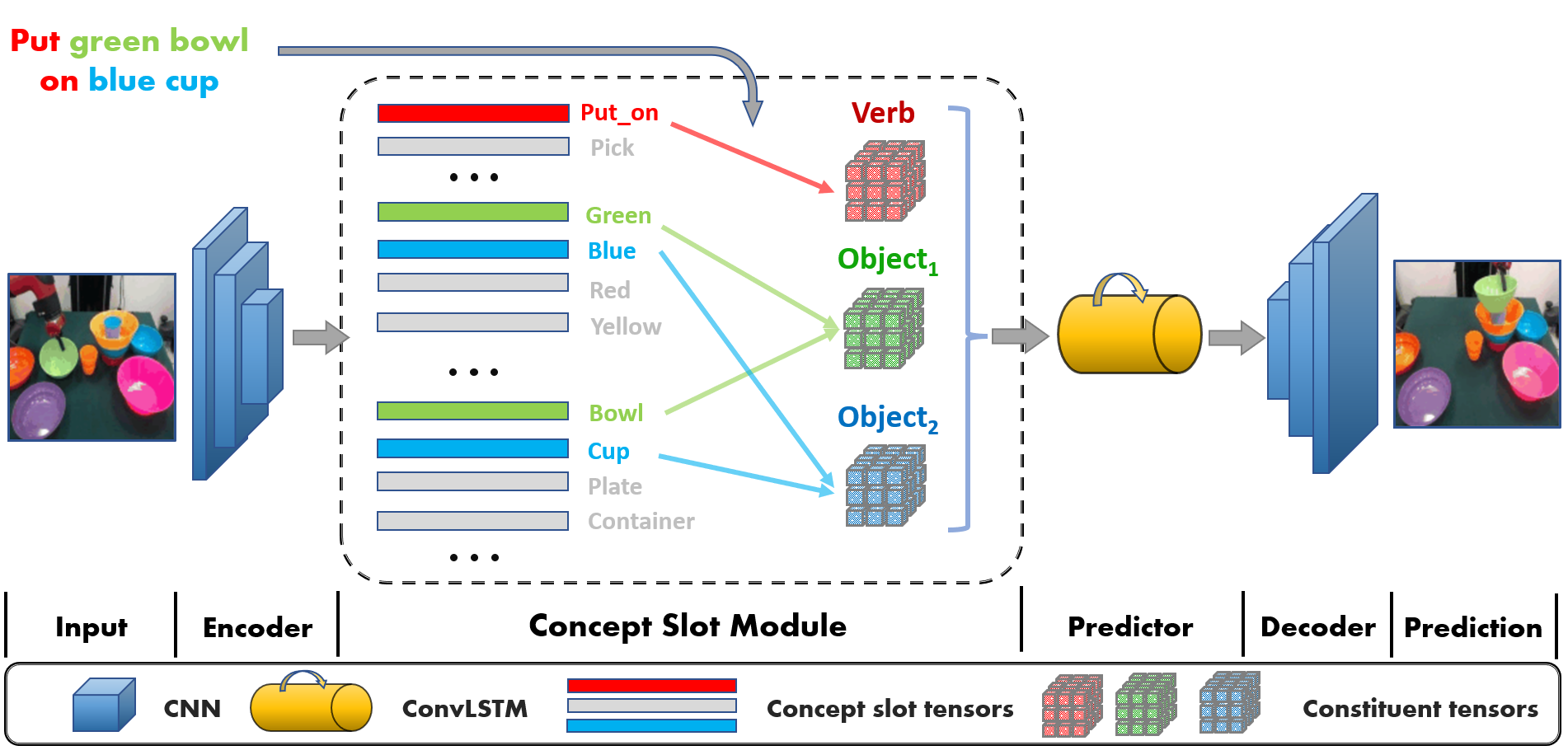}
		\caption{The pipeline of MAC in which the computation of concept slot module is elaborated (Better viewed in color). Feature maps extracted by encoder are mapped into the concept slot tensors. Concept slot module receives an action label that controls the collection of concept slot tensors and outputs representations encapsulating this action. A recurrent predictor updates representations before sending them to decoder to predict the next frame.}
		\label{figure:model}
		\vspace{-5pt}
\end{figure*}


\section{Approach}
\label{approach}
We begin with defining the task of semantic action-conditional video prediction. Given an initial frame $x_0$ and a sequence of action labels $a_{1:T}$, the model is required to predict the corresponding future frames ${x_{1:T}}$. Each action label is a pre-defined semantic description of a spatiotemporal movement that involves multiple objects in a scene and spans over multiple frames such as "\textit{take the yellow cup on the table}" from $t=0$ to $t=10$. So technically, one can regard this task as an inverse problem of action recognition. 
It should also be pointed out that our semantic task is different from common \textit{dense video prediction and generation tasks} in the sense that it focuses on predicting \textbf{time-agnostic} events. Hence, we design the corresponding datasets as videos capturing sufficient key frames of entire actions. In future practices, we can further apply video interpolation methods in CV or motion planner algorithms in RL to make up the intermediate process if needed. 

\subsection{Motivation}
The design of our new task is necessary for studying \textbf{compositional generalization} as it detaches the definition of object from its specific location. However, it also requires a successful model to figure out where the desired object is through leveraging abstract labels.  Our main idea is that we create a large number of small, specialized and relatively independent learners called concept slots for each word in the dictionary of action labels to capture their corresponding spatial representations from observations. During training, action labels will be 
translated as constituency trees to control the activations of all related concept slots and to assemble the representations of given actions for next-frame prediction. As a result, this language-guided gating mechanism embeds the syntactic structures into the learning system and enables the proposed model to dynamically recombine its learnt concepts so that it can understand the combinatorial complexity of the world. In this paper, we demonstrate that our method possesses many key characteristics of \textit{system-2} learning \cite{goyal2019recurrent,goyal2020inductive}, including concept grounding, sample efficiency, counterfactual generations, out-of-distribution generalization and fast transfer.


\subsection{Modular Action Concept Network}
The MAC model is composed of 4 modules including encoder $\mathcal{E}$, decoder $\mathcal{D}$, concept slot module $\mathcal{C}$ and recurrent predictor $\mathcal{P}$.
The goal of our model is to learn the following mapping:
\vspace{-1pt}
\begin{equation}
    \hat{x}_{t} = \mathcal{D}(\mathcal{P}(\mathcal{C}(\mathcal{E}(x_{t-1})|a_t)|h_{t-1}))
\end{equation}
where $x_t$, $a_t$ and $h_t$ are video frame, action labels and  hidden states at time $t$. The overall architecture of our method is illustrated in Fig \ref{figure:model}. In the case of stochastic video generation, another two  modules, prior $p(z)$ and posterior $q(z)$ , will be added to help estimate the latent distribution of trajectories.

\textbf{Encoder and Decoder}: At each timestep $t-1$, the encoder $\mathcal{E}$ receives visual input $x_{t-1}$ and extracts a set of multi-scale feature maps. 
In the deterministic setting, we employ a convolutional neural network with an architecture similar to DCGAN~\cite{radford2015unsupervised}.  The matching decoder $\mathcal{D}$ is a mirrored version of the encoder with down-sampling operations replaced with spatial up-sampling and additional sigmoid output layer. It  aggregates the updated latent representations produced by predictor and multi-scale feature maps from encoder to predict the next frame $\hat{x}_{t}$. 

In the stochastic setting, we use invertible autoencoder introduced in CrevNet~\cite{yu2019efficient} instead as we find this information-preserving architecture can better preserve the attributes of randomly moving objects. The corresponding decoder is the backward pass, i.e. inverse computation, of the same network of the encoder. Readers can find more details about invertible autoencoder and coupling layer in Appendix B.

\textbf{Concept Slot Module}: The concept slot module $\mathcal{C}$ is the core module of MAC. It resembles the mixture of experts as each slot focuses on only one concept in the space of action labels and will be activated and assembled  to represent the given actions through the language-guided gating functions. 

 Each atomic action label will first be decomposed into several constituents of sentence. A constituent is a verb or object phrase, like  “\textit{pick}” or "\textit{large red bowl}". Since we are mostly dealing with manipulation videos, atomic actions are usually divided into 3 constituents, verb, $\text{object}_1$, $\text{object}_2$, and more complex multiple-entity actions can be expanded into temporal sequences of several atomic two-object actions. For single-entity actions, $\text{object}_2$ will be filled with all zero tensors. Each constituent will have its own dictionary recording all pre-defined words or concepts and gating functions can be derived based on these dictionaries to establish bottom-up connections from concept slots. The computation of concept slot module is given as follows:
\begin{align}
    \textbf{w}^i =\Psi^i(\textbf{f}),
    \textbf{c}^j =\Phi^j(\text{Concat}(\{\textbf{w}^i|~\forall i,~ \delta^j(i)=1\}))
\end{align}
 where \textbf{w} and $\textbf{c}$ are concept and constituent representations and $\delta^j$ is the indicator function for gating function of $j_{\text{th}}$ constituent. More specifically, after the feature maps \textbf{f} are extracted from the input image, they are fed into $\mathcal{K}$ convolutional units  $\Psi^i$, i.e. the concept slot layer, to create $\mathcal{K}$ concept slot tensors of dimension $N_d$. Here, $\mathcal{K}$ is the total number of possible concepts we pre-defined in the dictionary of action labels. Since verbs can be interpreted as spatiotemporal changes of relationships between objects, not only slots for objects but also slots for verbs, like '\textit{take}' or '\textit{put on}', are computed from the extracted feature maps. 
 
 Next, a gating function will collect all involved concept slot tensors and create an ensemble as input for each constituent. This assembly process simulates the formation of simplified constituency parse trees. Constituent slot layer $\Phi^j$ can either be resolution-preserving or upsampling operators as spatial information is important for our new task. Finally, outputs of all constituent slots are concatenated pixel-wisely to obtain the representation of actions before sending them to predictor. It is worth noticing that MAC is allowed to have multiple concurrent actions in a scene at inference time. In this case, we copy additional groups of trained constituent slots to represent other actions.

\begin{figure}[!t]
	\centering
	\includegraphics[width=1.0\textwidth]{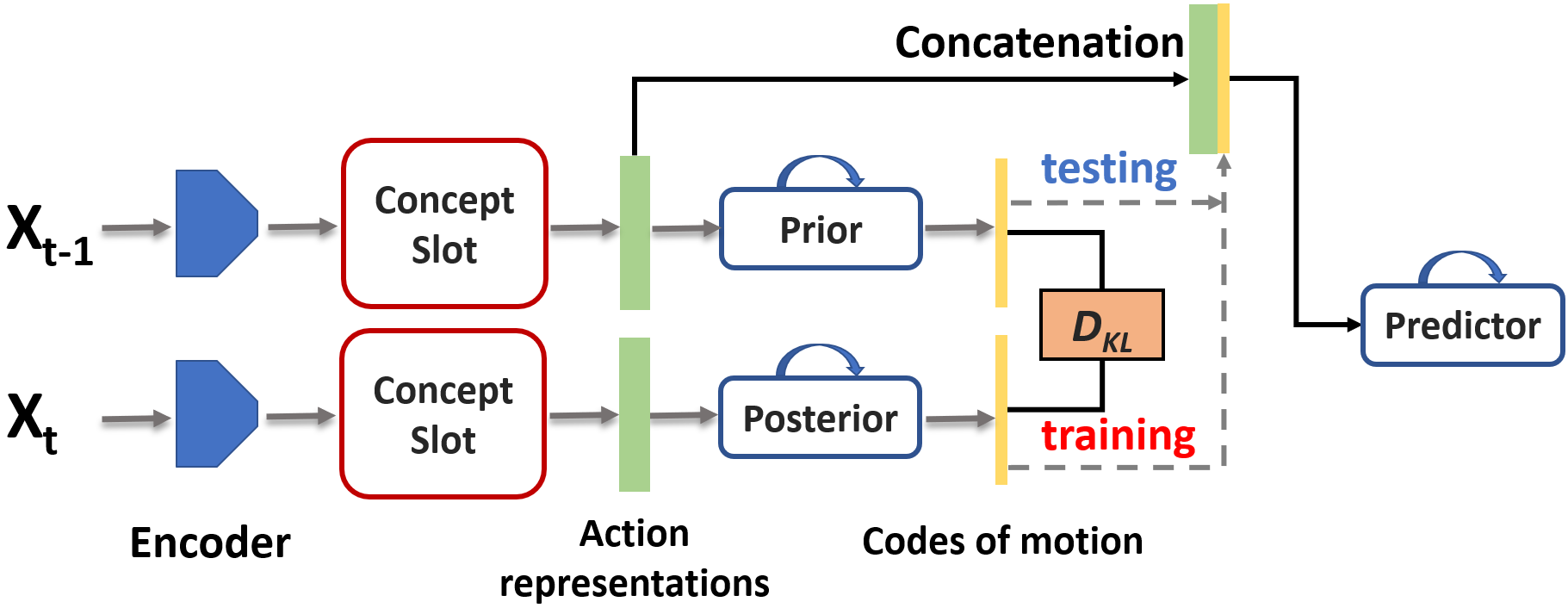}
	\caption{\textit{Learned prior}. Two recurrent inference modules are deployed to estimate the latent distribution of trajectories. The posterior inference network $q(z)$
can access to the representations of target frames to estimate a true distribution that we expect its prior counterpart $p(z)$  to mimic at test time. }
		\label{figure:lp}
\end{figure}

 \textbf{Learned Prior}: We leverage a technique called \textit{learned prior} (Fig \ref{figure:lp}) from SVG~\cite{denton2018stochastic} to model the stochastic movements in videos. In particular, we build two additional recurrent inference networks, prior and posterior respectively, to capture the randomness of motions. During training, the posterior inference network $q(z)$ can access to the representations of target frames to estimate a true distribution of trajectory that we expect its prior counterpart $p(z)$  to mimic at test time. Codes of motions $z_{t}$ estimated by posterior during training (or by prior during testing) will then be concatenated with latent representations before sent to predictor.  

\textbf{Predictor}: The recurrent predictor $\mathcal{P}$, implemented as a stack of residual ConvLSTM layers \cite{xingjian2015convolutional}, calculates the spatiotemporal evolution for each action label respectively. The memory mechanism of ConvLSTM is essential for MAC to remember its previous actions and to recover the occluded objects. To prevent interference between concurrent actions, hidden states are not shared between actions. The outputs of predictor for all action labels are added point-wisely.

\textbf{Training}: In the deterministic setting, we train our model by minimizing the mean squared error the between the target frames and the predictions. In the stochastic setting, we optimize the following variational lower bound (ELBO) using re-parameterization trick \cite{kingma2013auto}:
\vspace{-10pt}
\begin{align}
 \mathcal{L}_{\theta,\phi,\psi}(x_{1:T}) = & \sum_{t=1}^{T} [\mathbb{E}_{q_{\phi}(z_{1:t}|x_{1:t})}\log p_\theta(x_t|z_{1:t},x_{1:t-1}) \nonumber  \\
 & -\beta D_{KL}(q_\phi(z_t|x_{1:t})||p_\psi(z_t|x_{1:t-1}) ] 
\end{align}
where $p_\theta$ is the future frame generator, $z_t$ represents the latent codes of motion, $p_\psi(z_t|x_{1:t-1})$ is the prior distribution, $q_\phi(z_t|x_{1:t})$  is the posterior distribution and $D_{KL}$ denotes the Kullback–Leibler (KL) divergence which forces the posterior to approximate the prior distribution. Since $p_\theta$ is modeled by conditional Gaussian, the likelihood term reduces to MSE measure between the ground truth frames and the predictions. The full derivation of ELBO is provided in the Appendix A. 

At the inference phase, the model will use its previous predictions as visual inputs instead except for the first pass. Hence, a training strategy called scheduled sampling~\cite{bengio2015scheduled} is adopted to alleviate the discrepancy between training and inference. 

\section{Datasets}
In this study, we create two new synthetic datesets, CLEVR-Building-blocks and Sapien-Kitchen, and label one real-world dataset called Tower-Creation from Roboturk \cite{mandlekar2018roboturk} for evaluation. This is because most existing video datasets either don't come with semantic action labels~\cite{babaeizadeh2017stochastic} or fail to provide necessary visual information in their first frames due to egomotions and occlusions~\cite{hundt2018costar}. Although there are several candidate datasets like Penn Action~\cite{zhang2013actemes}, BAIR \cite{finn2016unsupervised} and KTH~\cite{schuldt2004recognizing} for multi-modal learning, they all adopt the same single-entity setting which actually indicates they can be solved by a much simpler model. To tackle the above issues, we design each video in our datatsets as a depiction of certain atomic action performed by an agent with objects which are observable in the starting frame. Furthermore, we add functions to generate bounding boxes of all objects for both synthetic datasets in order to train AG2Vid. It is worth noting that all three of these domains exhibit a key property named \textbf{combinatorial explosion}, resulting in factorial complexity growth in both spatial and temporal dimensions even with a small object set. For instance, a sequence with 6 (out of 32) objects and 6 actions can have 333,396,000 possibilities without considering any continuous factor. Hence, our model only sees a small fraction of these potential scenarios during training. 




\subsection{CLEVR-Building-blocks Dataset}
CLEVR-Building-blocks dataset is built upon CLEVR environment~\cite{johnson2017clevr}. For each video, the data generator  initializes the scene with 4 - 6 randomly positioned  and visually different objects. There are totally 32 combinations of shapes, colors and materials of objects and at most one instance of each combination is allowed to appear in a video sequence. The agent can perform one of the following 8 actions on objects $\mathcal{O}_A$ and $\mathcal{O}_B$: \textit{Pick $\mathcal{O}_A$}, \textit{Pick and Rotate $\mathcal{O}_A$ transversely / longitudinally}, \textit{Put $\mathcal{O}_A$ on $\mathcal{O}_B$}, \textit{Put $\mathcal{O}_A$ on the left / right side of $\mathcal{O}_B$}, \textit{Put $\mathcal{O}_A$ in the front of / behind $\mathcal{O}_B$}. Each training sample contains a video of three consecutive \textit{Pick-} and \textit{Put-} action pairs and a sequence of semantic action labels of every frame.



\subsection{Sapien-Kitchen Dataset}
Sapien-Kitchen Dataset describes a more complicated environment in the sense that: (a). It contains deformable actions like \textit{"open"} and \textit{"close"}; (b). The structures of different objects in the same category are highly diverse; (c). Objects can be initialized with randomly assigned relative positions like \textit{"along the wall"} and \textit{"on the dishwasher"}. We collect totally 21 types of small movable objects in 3 categories, \textit{bottle, kettle} and \textit{kitchen pot}, and 19 types of large openable appliances in another 3 categories, \textit{oven, refrigerator} and \textit{dishwasher}, from Sapien engine~\cite{xiang2020sapien}. The agent can perform one of the following 6 atomic actions on small object $\mathcal{O}_s$ and large appliance $\mathcal{O}_l$: \textit{Take $\mathcal{O}_s$ on $\mathcal{O}_l$}, \textit{Take $\mathcal{O}_s$ in $\mathcal{O}_l$}, \textit{Put $\mathcal{O}_s$ on $\mathcal{O}_l$}, \textit{Put $\mathcal{O}_s$ in $\mathcal{O}_l$}, \textit{Open $\mathcal{O}_l$} and \textit{Close $\mathcal{O}_l$}. Composite action sequences are defined as follows: \textit{"Take\_on--Put\_on"}, \textit{"Take\_on--Open--Put\_in--Close"}, \textit{"Open--Take\_in--Close"}. 

\begin{table*}[!t]
\vspace{-15pt}
	    \footnotesize
			\begin{tabular}{ p{3.1cm}p{0.8cm}p{0.7cm}p{0.9cm}p{1.2cm}|p{0.8cm}p{0.7cm}p{0.9cm}p{1.1cm} }
			\toprule
				\multirow{2}{*}{Model}  & \multicolumn{4}{c|}{CLEVR-Building-blocks}  & \multicolumn{4}{c}{Sapien-Kitchen}\\
				
				& SSIM$\uparrow$ & MSE$\downarrow$ & LPIPS$\downarrow$  &  Accuracy$\uparrow$ & SSIM$\uparrow$ & MSE$\downarrow$ & LPIPS$\downarrow$  &  Accuracy$\uparrow$\\
			\midrule
			 Copy-First-Frame & 0.962  & 251.38 & 0.1320  & - & 0.951 & 152.87 & 0.0393 & - \\
			 \midrule
			 Concatenation Baseline & 0.961  & 226.53 & 0.1301  & 50.8\% & 0.962 & 23.13 & 0.0232 & 52.4\% \\
			 AG2Vid  & 0.956  & 58.67 & 0.0399  & 78.8\%  & 0.947 & 270.87 & 0.0684 & 5.2\% \\
			 MAC & \textbf{0.983}  & \textbf{43.52} & \textbf{0.0303}  & \textbf{95.2\%} & \textbf{0.971} & \textbf{11.16} & \textbf{0.0178} & \textbf{86.4\%} \\
			\bottomrule	
			\end{tabular}
    \caption{Quantitative evaluation on CLEVR-Building-blocks and Sapien-Kitchen. All metrics are averaged frame-wisely except for accuracy.} \label{table:quant}
\end{table*}

\subsection{Tower-Creation Dataset}
Each video in Tower-Creation Dataset depicts a robotic arm building a tower with flatware present on the table. We have labeled 524 videos in total since semantic descriptions are not provided and produce 1867 samples consists of two actions: \textit{Pick $\mathcal{O}_A$} and \textit{Put $\mathcal{O}_A$ on $\mathcal{O}_B$}. We use 1536 video clips for training and 331 for evaluation. It should be pointed out that the size of Tower-Creation dataset is small compared with commonly used datasets such as BAIR \cite{finn2016unsupervised}  which has 59k videos in total. Thus, our experiments can also tell whether evaluated methods are data efficient.  

\section{Experimental Evaluation}
\begin{figure}[!t]
	\centering
		\includegraphics[width=1.0\textwidth]{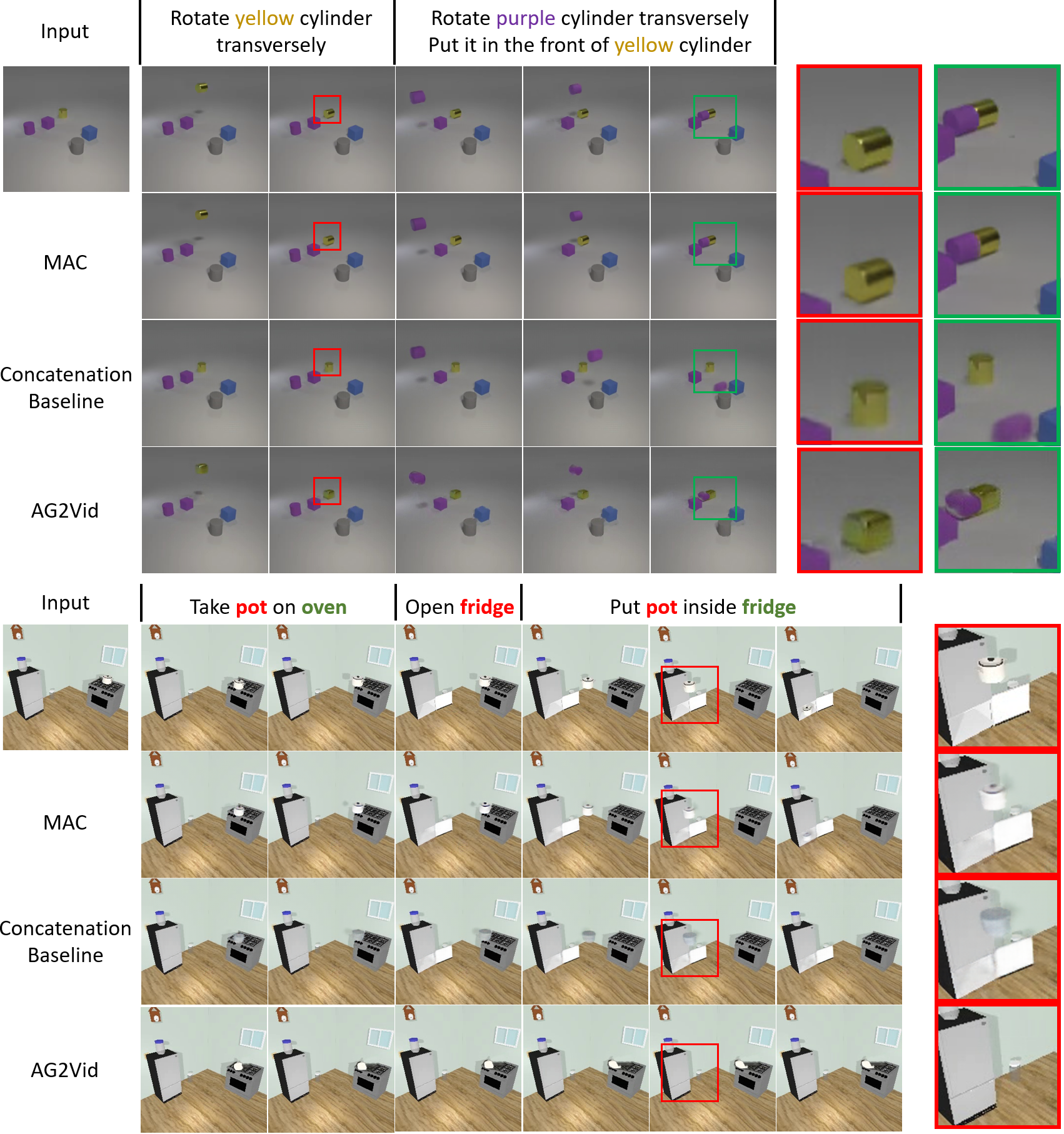}
		\caption{The qualitative comparison on CLEVR-Building-blocks and Sapien-Kitchen. The first row of each figure is the groundtruth sequence. The red and green boxes highlight the quality of predictions by each method. In contrast to the success of MAC, concatenation-based method fails to find the correct destinations or to preserve attributes of moving objects. Also, bounding boxes used in AG2Vid  cannot portray visual changes like rotations correctly.}
		\label{figure:qual}
\end{figure}

\begin{figure*}
\floatbox[{\capbeside\thisfloatsetup{capbesideposition={right},capbesidewidth=3.5cm}}]{figure}[\FBwidth]
{\caption{  \textbf{Counterfactual video generation}: Conditioning on the same initial frame and different action labels, MAC can produce high-quality imaginations of counterfactual futures. Various visual outcomes present in the final frames are highlighted with red boxes and enlarged in the final column.\\
\textbf{Top}: Generative results on CLEVR-Building-blocks. 34  frames are generated.\\
\textbf{Bottom}: Generative results on Sapien-Kitchen dataset. 35 frames are generated.} \label{figure:counterfactual}}
{\includegraphics[width=13.25cm]{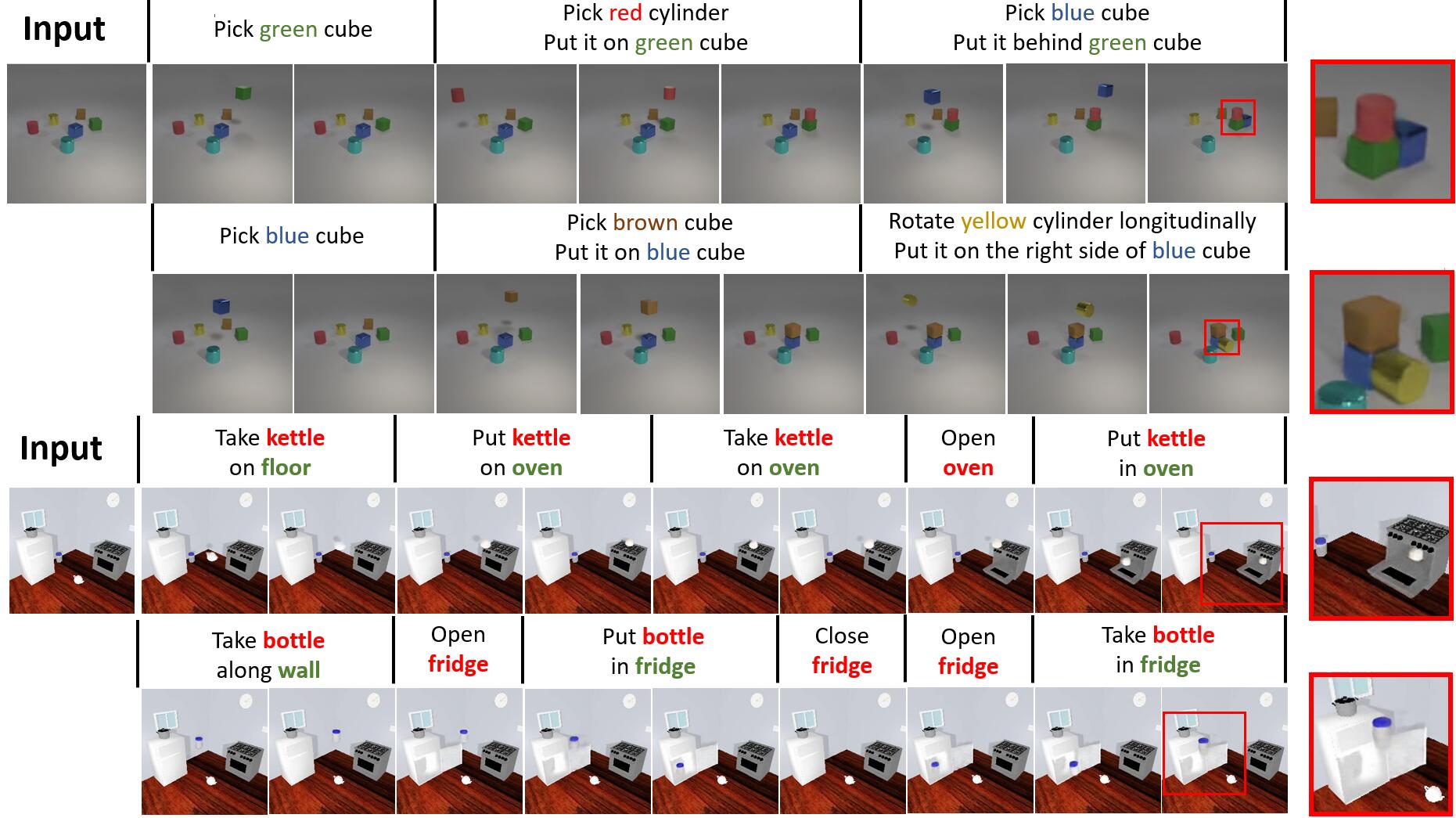}}
\end{figure*}

\subsection{Action-conditonal video prediction}
\textbf{Baselines and setup}: We evaluate the proposed model on CLEVR-Building-blocks and Sapien-Kitchen Datasets. AG2Vid~\cite{bar2020compositional} is re-implemented  as the baseline model because it is the most related work. Every action graph used in AG2Vid can be equivalently translated into our cases because each atomic action graph in AG2Vd also involves at most two objects. But unlike our method which only needs visual input and action sequence, AG2Vid also requires bounding boxes of all objects for training and testing and it can only handle deterministic prediction. 
Furthermore, we conduct an ablation study by replacing concept slot module with the concatenation of features and tiled action vector, which is commonly used in low-level action-conditional video prediction~\cite{finn2016unsupervised}, to show the effectiveness of our module. 


\textbf{Metrics}: To estimate the fidelity of action-conditional video prediction, MSE, SSIM~\cite{wang2004image}, PSNR and  LPIPS~\cite{zhang2018unreasonable} are calculated between the predictions and groundtruths. FID \cite{heusel2017gans} and FVD \cite{unterthiner2018towards} are not appropriate for this task because they cannot tell how faithfully the model obeys the given instructions. However, these metrics still may not effectively tell if actions are successfully completed due to the small sizes of the moving objects. Hence, we also perform a human study to assess the accuracy of performing the correct action in generated videos for each model. The human judges annotate whether the model can identify the desired objects, perform actions specified by action labels and maintain the consistent visual appearances of all objects in its generations and only videos meeting all three criterions are scored as correct. Also, we find it is technically infeasible to train an action recognition model to estimate the accuracy due to the innumerable action labels caused by the property of combinatorial explosion.

\textbf{Results}\label{result}:
The quantitative comparisons of all methods are summarized in Table 1. The MAC achieves the best scores on all metrics without access to additional information like bounding boxes, showing the superior performance of our concept slot module. The qualitative analysis in Fig \ref{figure:qual} further reveals the drawbacks of other baselines. For CLEVR-Building-blocks, the concatenation-based variant fails to recognize the right objects due to its limited inductive bias. Although AG2Vid has no difficulty in identifying the desired objects, assumptions made by flow warping are too strong to handle rotation and occlusion. Consequently, the adversarial loss enforces AG2Vid to fix these errors by converting them to wrong poses or colors. These limitations of AG2Vid will be further amplified in a more complicated environment, i.e. Sapien-Kitchen. The same architecture used for CLEVR can only learn to remove the moving objects from their starting positions in Sapien-Kitchen because rotation and occlusion occur more often. The concatenation baseline performs better by showing correct generation of open and close actions on large appliance. Yet, it still fails to produce long-term consistent predictions as the visual appearances of moving objects are altered. On the contrary, MAC can authentically depict the correct actions specified by action labels on both datasets.

\subsection{Counterfactual generation}
\textbf{Counterfactual generation}: The most intriguing application of MAC is counterfactual generation. More specifically, counterfactual generation means that our model will observe the same starting frame but receive different valid action labels to produce the corresponding future frames.


 \begin{figure*}
\begin{tabular}{cc}
    \begin{minipage}{0.66\textwidth} \includegraphics[width=1.0\textwidth]{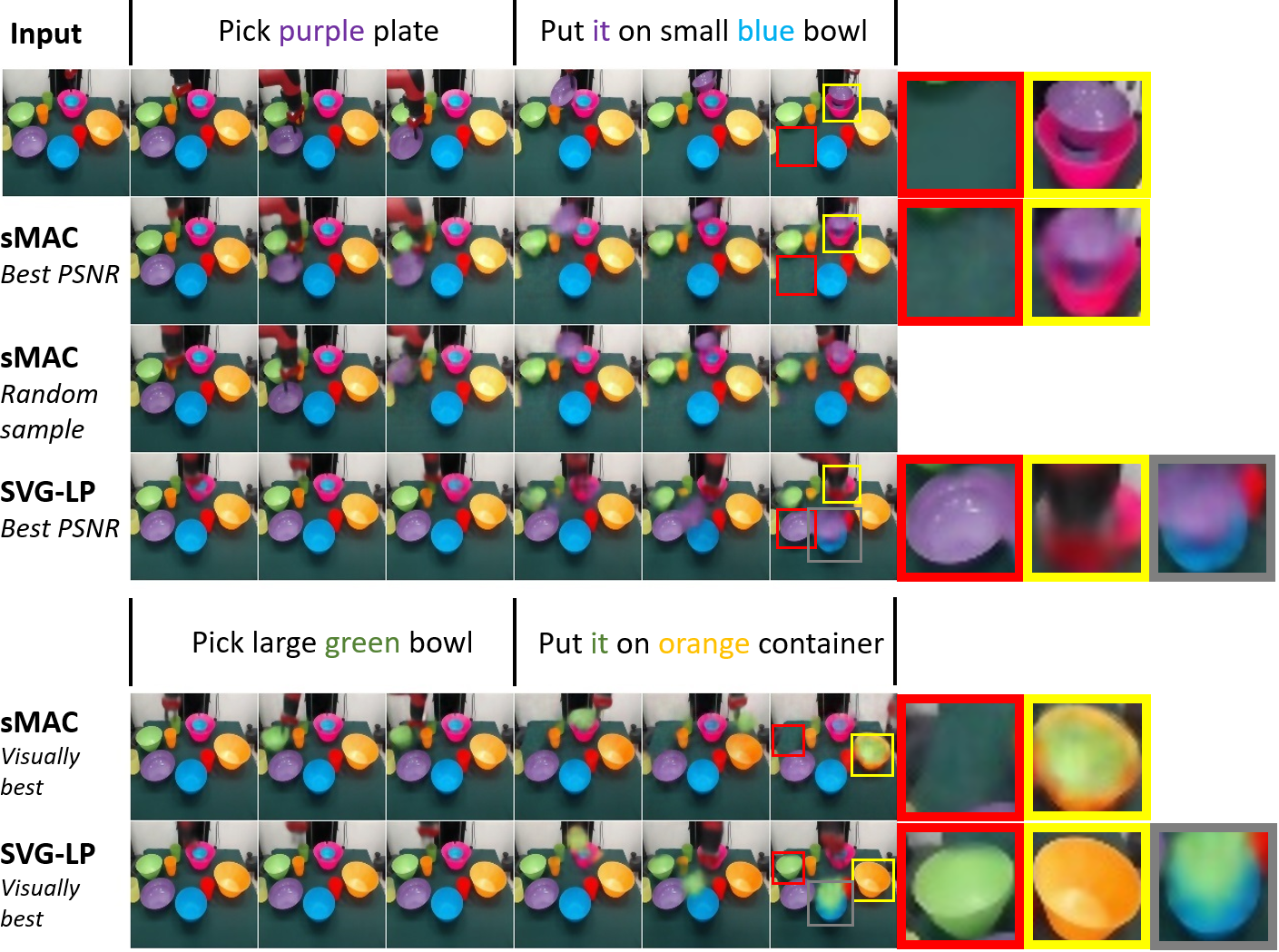} \end{minipage}& \begin{minipage}{0.3\textwidth} \includegraphics[width=1.15\textwidth]{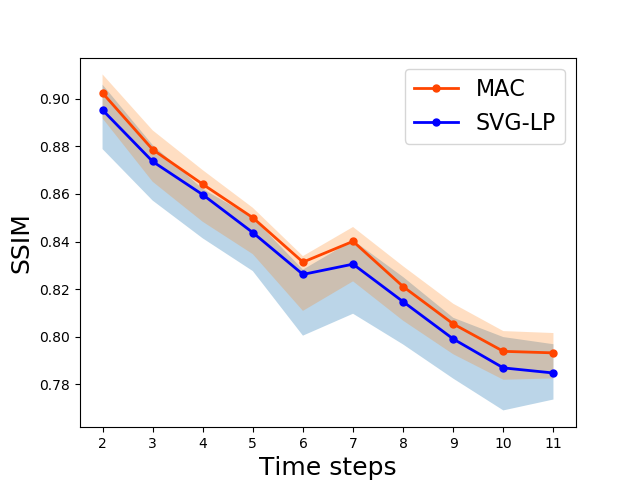} \\ \includegraphics[width=1.15\textwidth]{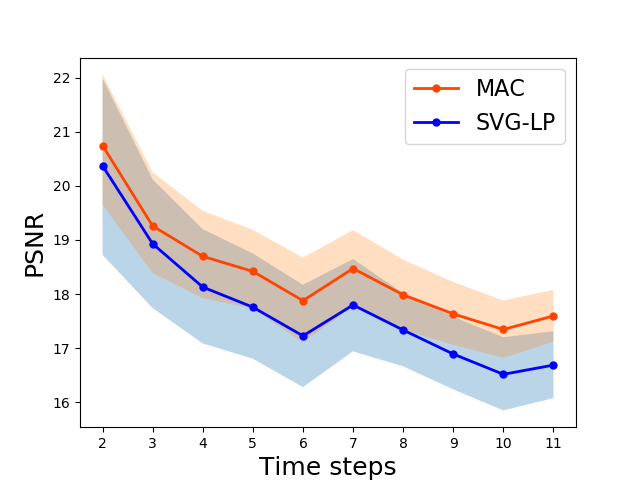} \end{minipage}
    \end{tabular}
        \caption{\footnotesize \textbf{Left}: Visual comparison between sMAC and SVG-LP on Tower-Creation.  The supposed completions of \textit{Pick} and \textit{Put} in the final frames are highlighted by red and yellow boxes while incorrect completions in SVG-LP generations are labelled by  grey boxes. The last two rows are counterfactual generations in which models are given different action labels. \textbf{Right}: Quantitative comparison per-frame. Higher SSIM and PSNR indicate better performance. \label{figure:svg}}
    \end{figure*}
\textbf{Results}:  The visual results of counterfactual generations on each dataset are displayed in Fig \ref{figure:counterfactual}. As we can see, our model successfully identifies the desired objects, plans correct trajectories toward the target places and generates high-quality imaginations of counterfactual futures. It is also worth noticing that all displayed generations are long-term generations , i.e. more than 30 frames are predicted for each sequence. Our recurrent predictor plays an very important role in sustaining the spatiotemporal consistency and in reconstructing the fully-occluded objects.

\subsection{Stochastic video generation}
\textbf{Baselines and setup}: We continue to evaluate the stochastic version of MAC (sMAC) on Tower-Creation dataset. SVG-LP was extended to action-conditional version in \cite{villegas2019high} so that we can adopt it as the baseline model to demonstrate the effectiveness of concept slot module. 

\textbf{Results}: The qualitative and frame-wise quantitative comparison between sMAC and action-conditional SVG-LP is provided in Fig \ref{figure:svg}. Although SVG-LP can  partially understand the given action labels, it often fails to locate and manipulate the desired objects. Consequently, it will generate the moving object out of nowhere and often place it on a wrong target object. In contrast, sMAC can successfully simulate the trajectory of robotic arms and correctly animate the "\textit{Pick}" and "\textit{Put}" actions thanks to the concept slot module. Row 3 and 5 in Fig \ref{figure:svg} show that sMAC is also capable of producing diverse future frames and predicting counterfactual results following different action instructions. The overall accuracy of sMAC estimated by human study on Tower-Creation is 65.3\% compared with 31.8\% of SVG-LP.

\begin{figure*}
\floatbox[{\capbeside\thisfloatsetup{capbesideposition={right},capbesidewidth=3.5cm}}]{figure}[\FBwidth]
{\caption{ \footnotesize  Compositional generalization and feature reuse.\\
\textbf{Top}: Unobserved scenarios. All red cubes are removed from the tranining data, but the trained model can still manipulate red cube at test time.\\ 
\textbf{Middle}: Concurrent actions. Inputting two action sequences at the same time. Both actions are depicted correctly.\\
\textbf{Bottom Left}: New-object adaptation. Even with a few training samples, MAC can be fast adapted for generation of new objects. Red arrows point to new objects present in images\\
\textbf{Bottom Right}: Object detection. Learnt features can be directly applied for detection. 
}\label{figure:ood}}
{\includegraphics[width=13.25cm]{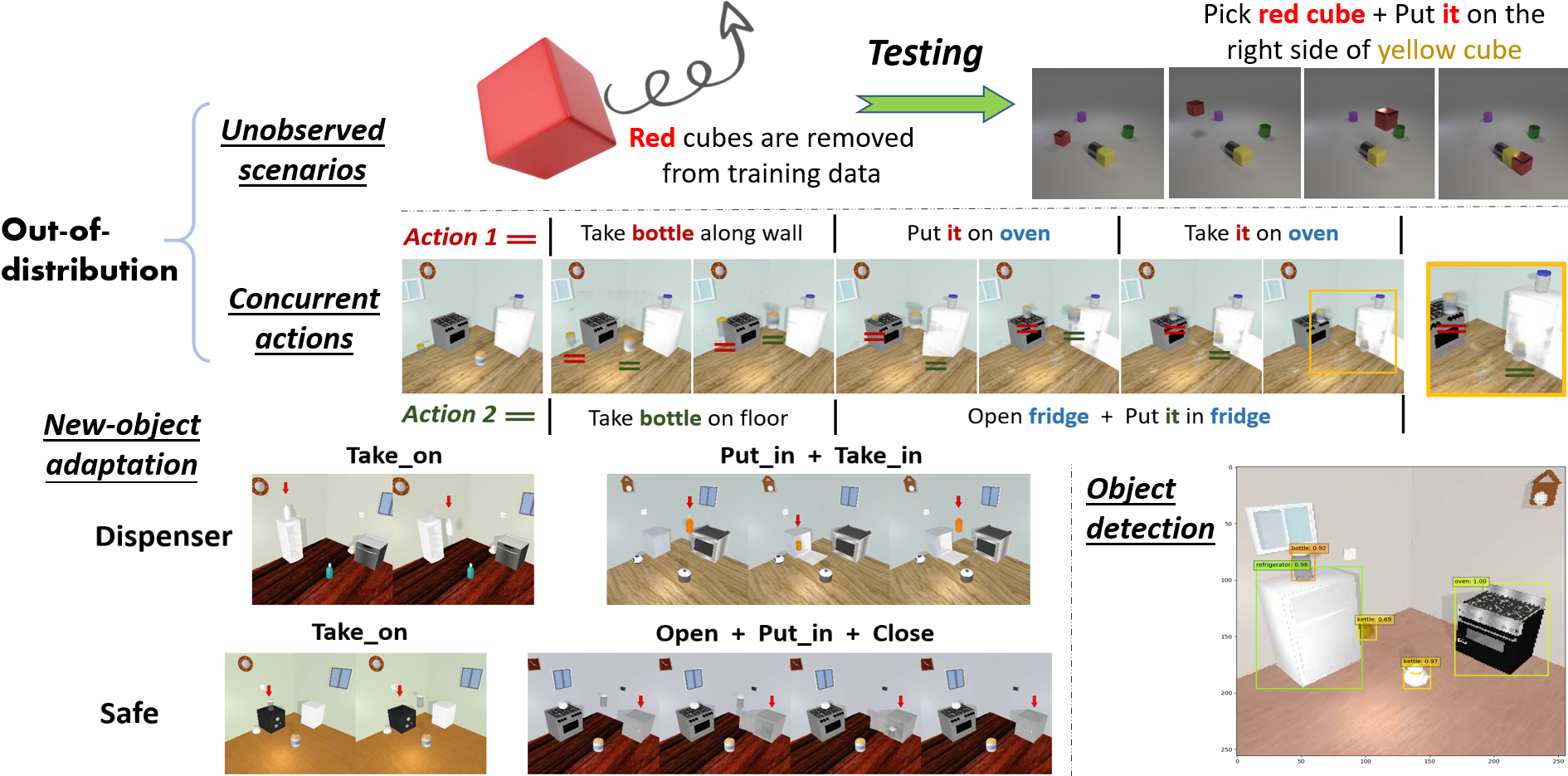}}
\end{figure*}

\subsection{Compositional Generalization}
We further explore other interesting features of our MAC. We first demonstrate that  MAC is capable of making out-of-distribution generalization by designing two experiments. We evaluate how quickly our model can be adapted to new objects. It turns out for each new object, the trained MAC only requires a few training video examples to generate decent results. Finally, to verify that our model encodes the spatial information, we add SSD~\cite{liu2016ssd} head after the frozen encoder and concept slot layer to conduct object detection.

\textbf{Unobserved scenarios}: We design an interesting experiment where only a subset of CLEVR-Building-blocks data are used for training and check what will happen if we input the unobserved action labels to the trained model. More precisely, we exclude all videos manipulating red cubes in the training sets and send the instructions involving red cubes at test time. Note that we only remove one object to avoid high correlation among concept slots, otherwise it will violate the relative independence assumption. Since failure cases will not manipulate the correct objects and will produce very large pixelwise losses. We can set a threshold of MSE to calculate the accuracy of performing the correct actions, which is 75.6\%. The visualization of this experiment can be found in Fig \ref{figure:ood}. As we can see, MAC can still identify and manipulate red cubes correctly, showing its ability to recombine the learnt concept to comprehend new objects.

\textbf{Concurrent actions}: Concurrent actions means multiple action inputs at the same time. It can be considered as out-of-distribution generalization because our model only observes single-action videos during training. 
Generating concurrent-action videos needs to employ copied constituent slots and parallel hidden states. As illustrated in Fig \ref{figure:ood}, MAC can linearly integrate the action information in the latent space and correctly portray 2 concurrent actions in the same scene.


\textbf{Adaptation}: We add a new openable category "\textit{safe}" and a new movable category "\textit{dispenser}" into Sapien-Kitchen and generate 100 video sequences for each new object showing its interaction with other objects. Approximately, there are about 5 new  sequences created for each new action pair between 2 objects. Blank concept slots for new categories are attached to trained MAC and we finetune it on this small new training set. Visualization in Fig \ref{figure:ood} shows that even with a few training samples,  MAC is accurately adapted for video generation of new objects. This is because, with the help of concept slots, MAC can disentangle actions into relatively independent grounded concepts. When it learns new concepts, MAC reuses and integrates prior knowledge learnt from different cases.

\textbf{Object detection}: The quantitative results of object detection and more visualizations can be found in Appendix D. We observe that the features learnt by MAC can be easily transferred for detection as our video prediction task is highly location-dependent. This result indicates that utilizing bounding boxes might be a little redundant for some video tasks because videos already provide rich motion information that can be used for salient object detection.



	

\section{Related Work}
\textbf{Video prediction}: ConvLSTM~\cite{xingjian2015convolutional} was the first deep learning model that employed a hybrid of convolutional and recurrent units for passive video prediction. This architectural design was soon followed by studies looking at a similar problem~\cite{mathieu2015deep,wang2017predrnn, yu2019efficient,wang2018eidetic}. However, the capability of passive video prediction framework is very limited as models usually don't have sufficient information to predict the long-term future due to partial observation, egomotion and randomness. It also prevents models from interacting with  environment.

On the other hand, the low-level action-conditional video prediction task provides an action vector at each timestep as additional input to guide the prediction~\cite{oh2015action,chiappa2017recurrent,babaeizadeh2017stochastic,wu2021greedy}. CDNA~\cite{finn2016unsupervised} is a representative of such models. In CDNA, the states and action vectors of the robotic manipulator are first spatially tiled and integrated into the model through concatenation. SVG~\cite{denton2018stochastic} was initially proposed for stochastic video generation but later was extended to action-conditional version in~\cite{villegas2019high}. SVG also used concatenation to incorporate action information. Such implementations are prevalent in low-level prediction because the action vector only encodes the spatial information of a single entity, usually a robotic manipulator~\cite{finn2016unsupervised} or a human hand. A common failure case for such models is the presence of multiple entities~\cite{kim2019unsupervised}, a scenario that our task definition and datasets focus on.


\textbf{Modularity}: Mixture of Experts refers to a classical machine learning technique where various learners are employed, each of which specializes in one particular function, and their output are aggregated through a gating function. This modular design makes each submodule relatively independent and thus leads to better generalization and robustness to compositional changes, which has been studied in several works~\cite{goyal2019recurrent,afshar2021mixcaps,sabour2017dynamic,henaff2016tracking}. In this work, we hypothesis that the underlying syntactic structures of semantic labels can tell how to aggregate the representations of individual concept learners. By translating labels into constituency trees, action graphs are embedded into the learning system to get the entire perspective of ongoing activities while each concept learner can focus on its specific subtask.
\section{Limitations}
While the results of MAC are very impressive, there are still several limitations to this work, including (a). \textit{Uniqueness}: We didn’t design specific mechanisms that enable MAC to randomly select one of repeated objects. We assume each object is unique in the scene. (b). \textit{More flexible semantic instructions}: In this work, we use semantic labels pre-defined in a relatively fixed format. Therefore, we can translate each label to a constituency tree without using any learnable function.(c). \textit{Ego-motion}: All videos we evaluated on were captured by fixed cameras. Videos with ego-motions can provide a more abundant source of training data. 
\section{Conclusion}
In this work, we propose the new task of semantic action-conditional video prediction and introduce 3 new datasets that are meant to bridge the gap towards a robust solution to this task in complex interactive scenarios. MAC, a novel video prediction model, was also designed by utilizing the idea of MoE to ground action concept for video generation. Our proposed model can generate alternative futures without requiring additional auxiliary data such as bounding boxes, and is shown to be both quickly extendible and adaptable to novel scenarios and entities. 
It is our hope that our contributions will advance progress and understanding within this new task space, and that a model robust enough for real-world applications (i.e. in robotic systems) in control will be eventually proposed as a descendant of this work.

\noindent \textbf{Acknowledgement} ~ This work is supported by CIFAR AI Chair, NSERC Discovery Award, University of Toronto XSeed award, and gifts from LG. 
\newpage

{\small
\bibliography{iccv2}
\bibliographystyle{iclr2022_conference}
}
\end{document}